\documentclass[10pt,twocolumn,letterpaper]{article}

\usepackage{iccv}
\usepackage{times}
\usepackage{epsfig}
\usepackage{graphicx}
\usepackage{amsmath}
\usepackage{amssymb}

\usepackage[normalem]{ulem}
\useunder{\uline}{\ul}{}
\usepackage{float}
\usepackage[numbers]{natbib}

\usepackage[pagebackref=true,breaklinks=true,letterpaper=true,colorlinks,bookmarks=false]{hyperref}

\iccvfinalcopy 

\def\iccvPaperID{10173} 

\ificcvfinal\fi

\begin{document}
\title{ChildGAN: Large Scale Synthetic Child Facial Data Using Domain Adaptation in StyleGAN}

\author{Muhammad Ali Farooq\\
University of Galway\\
{\tt\small muhammadali.farooq@nuigalway.ie}
\and
Wang Yao\\
University of Galway\\
{\tt\small w.yao2@nuigalway.ie}
\and
Gabriel Costache\\
Xperi Corporation, Galway\\
{\tt\small gabriel.costache@xperi.com}
\and
Peter Corcoran\\
University of Galway\\
{\tt\small peter.corcoran@nuigalway.ie}
}
\twocolumn[{%
\maketitle
\begin{figure}[H]
\hsize=\textwidth 
\centering
\includegraphics[width=1.8\linewidth]{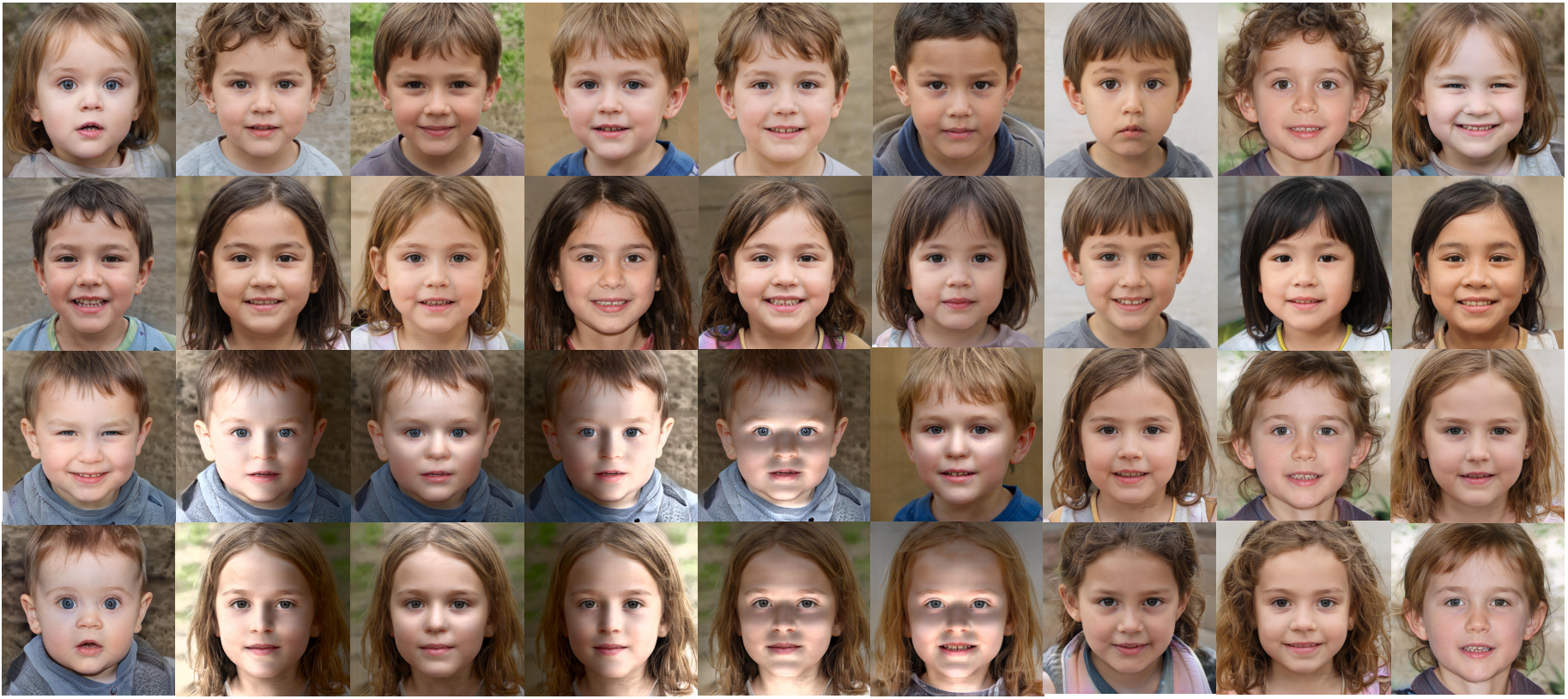}
\caption{Novel photo-realistic child data samples generated using ChildGAN with various smart image transformations.}
\label{fig12}
\end{figure}
}]

\begin{abstract}
In this research work, we proposed a novel ChildGAN, a pair of GAN networks for generating synthetic boys and girls facial data derived from StyleGAN2. ChildGAN is built by performing smooth domain transfer using transfer learning. It provides photo-realistic, high-quality data samples. A large-scale dataset is rendered with a variety of smart facial transformations: facial expressions, age progression, eye blink effects, head pose, skin and hair color variations, and variable lighting conditions. The dataset comprises more than 300k distinct data samples. Further, the uniqueness and characteristics of the rendered facial features are validated by running different computer vision application tests which include CNN-based child gender classifier, face localization and facial landmarks detection test, identity similarity evaluation using ArcFace, and lastly running eye detection and eye aspect ratio tests. The results demonstrate that synthetic child facial data of high quality offers an alternative to the cost and complexity of collecting a large-scale dataset from real children. 
\end{abstract}
\section{Introduction} \label{sec1}
\begin{table*}[htb]
\begin{center}
\resizebox{0.9\linewidth}{!}{
\begin{tabular}{|l|c|l|l|}
\hline
\multicolumn{1}{|c|}{Dataset} & \begin{tabular}[c]{@{}c@{}}Number of \\ Subjects\end{tabular} & \multicolumn{1}{c|}{Age} & \multicolumn{1}{c|}{Attributes} \\ \hline
\begin{tabular}[c]{@{}l@{}}The \\ DuckEES \\ Dataset~\cite{DuckEES} {}{}\end{tabular} &
  N/A &
  \begin{tabular}[c]{@{}l@{}}8 to\\ 18 \\ year \\ olds\end{tabular} &
  \begin{tabular}[c]{@{}l@{}}1. The DuckEES database contains dynamic emotional stimuli by children and adolescent actors.\\ 2. It has a greater representation of videos thus comprising 142 posed videos which are distributed\\ relatively homogeneously by emotion.\\ 3. However, the DuckEES dataset did not include some of the universal emotions such that anger,\\ contempt, surprise.\end{tabular} \\ \hline
\begin{tabular}[c]{@{}l@{}}The Child \\ Emotion \\ Facial \\ Expression   \\ dataset~\cite{ChildEmotiondb} {}{}\end{tabular} &
  \begin{tabular}[c]{@{}c@{}}132 \\ Children\end{tabular} &
  \begin{tabular}[c]{@{}l@{}}4 to\\ 6 \\ year \\ olds\end{tabular} &
  \begin{tabular}[c]{@{}l@{}}1. A total of 29 hours of video were recorded in the studio for capturing various child facial\\ expressions.\\ 2. The overall dataset comes with eight distinct facial expressions which include contempt, \\happiness, fear neutrality, disgust, anger, surprise, and sadness.\\ 3. The number of stimuli of each emotion is as follows: neutrality  87, happiness 363, disgust 170,\\ surprise 104, fear 153 (48 induced, sadness 144, anger 157, and contempt 183.\end{tabular} \\ \hline
\begin{tabular}[c]{@{}l@{}}The Child \\ Affective \\ Facial \\ Expression \\ (CAFE) \\ dataset~\cite{CAFE} {}{}\end{tabular} &
  \begin{tabular}[c]{@{}c@{}}154 Child \\ Models\end{tabular} &
  \begin{tabular}[c]{@{}l@{}}2 to\\ 8 \\ year \\ olds\end{tabular} &
  \begin{tabular}[c]{@{}l@{}}1. Comprises data of 7 different facial expressions which include sadness,  happiness, surprise,\\ anger, disgust, fear, and neutral face. \\ 2. Not all children were able to successfully pose for all 7 expressions, so all unsuccessful attempts\\ were eliminated from the set. \\ 3. The result output data consist of 1192 total photographs.\end{tabular} \\ \hline
\end{tabular}
}
\end{center}
\caption{Open-Source Child Datasets}
\label{tab1}
\end{table*}

The majority of artificial intelligence and machine learning applications require large, curated and diverse datasets. It can be challenging to perform optimal training and rigorous validation of deep learning networks with limited, imbalanced, noisy data and if the data has inherent bias due to limitations in the data acquisition process which may not be known at the time the data is collected. Data acquisition itself is an expensive, and time-consuming process, particularly when human subjects are involved, and in many cases, it is technically impossible or impractically expensive to accurately replicate acquisitions. Further, when data is collected from human subjects there is the additional complication of new data protection regulations for personally identifiable data.    

In the EU, the General Data Protection Regulations (GDPR) provides a broad interpretation of personally identifiable data (PID) which extends to facial biometrics and speech data. This creates significant complexity when gathering any video, image, or audio data of human subjects as the scope of use of such data and any subsequent data-processing must be clearly defined and explained to the data subject and explicit consent is normally required. Further, the data must be securely stored, and a range of rights must be supported, including the right of the data-subject to request the removal of the stored data at any time. This becomes even more complex when dealing with vulnerable data-subjects such as children where consent is required from legal guardians and ideally the child-subject should be informed about the scope of data use in non-technical language.   
Synthetic human facial data is not directly associated with a living human person and thus it is not subjected to data protection regulations. 

In this work we investigate the potential to adapt state-of-the-art (SOTA) generative neural models ~\cite{stylegan,stylegan2}, combining these with a range of other proven tools and algorithms to develop a large-scale synthetic dataset of facial biometrics for a vulnerable population – young children. This work is inspired by a need to adapt and tune a range of SOTA computer vision algorithms, trained on adult populations, for smart mobile and embedded vision systems that will engage with young children. The primary objective is to develop a cluster of GAN networks using StyleGAN2 ~\cite{stylegan2} to generate large-scale, gender-balanced, child synthetic facial data samples. To increase the utility of this data for a variety of AI applications it is important to introduce a broad range of smart transformations thus building new data attributes so that the base data samples can be adjusted to support numerous training and test use cases. The potential applications of this novel synthetic child dataset include but are not limited to child gender classification, face localization, facial landmark detection, child facial expression analysis, generating 3D facial poses using single or multiple 2D images, performing domain transfer to generate a variety of facial expressions, facial accessories, etc. The core contributions of this research study are as follows.

\begin{itemize}
\item The ChildGAN model is provided with tunings to enable separately generating boys and girls facial samples.
\item A large-scale high-quality synthetic dataset comprising more than 300k facial data samples is made available to the research community.
\item The dataset incorporates a variety of smart facial transformations such that 4 different facial expressions, age progression, eye blinking effects, head pose variations, skin and hair color variations, and distinct lighting conditions are available. 
\item The data quality is validated using various computer vision algorithms to test the uniqueness and facial features quality of rendered data.
\end{itemize}

\section{Literature review} \label{sec2}
Most of the publicly available large-scale face datasets such as FFHQ~\cite{stylegan}, CelebA~\cite{celeba}, VGGFace2~\cite{vggface2}, and DigiFace-1M~\cite{DigiFace} are usually acquired from adults. It is challenging to find large scale real child dataset with diversified data attributes. 
Table~\ref{tab1} summarizes the open-source child datasets that are publicly available to the best of our knowledge however the number of these datasets and the amount of information available through these datasets is significantly less as compared to large-scale adult face recognition (FR) datasets.
When coming towards generating synthetic data we can find various algorithms among which SMOTE~\cite{DeepSMOTE}, ADASYN~\cite{ADASYN}, Variational AutoEncoders~\cite{8285168}, and Generative Adversarial Networks (GANs)~\cite{gan} are a few techniques for generating precise results. In this research, we have focused on generating a scalable set of synthetic child data using GAN architecture. When producing large-scale synthetic data, it should be aimed that generated samples are statistically accurate and free of historical biases. This will eventually be beneficial for the optimal training and validation of dense machine learning pipelines also referred to as "black box"~\cite{tiwald2021representative}.  

With the recent advancements in designing more efficient deep learning models, we can find numerous SoA open-sourced GAN architectures based on different loss functions. Some of these include the standard Vanilla GAN~\cite{gan}, LSGAN~\cite{lsgan},  WGAN~\cite{wgan}, WGAN with gradient penalty~\cite{E-WACGAN}, Deep Regret Analytic GAN~\cite{kodali2017convergence,fedus2017many}, CramerGAN~\cite{bellemare2017cramer}, Conditional WGAN~\cite{ZHENG20201009}, and GAN for time series as well also known as TimeGAN~\cite{NEURIPS2019_c9efe5f2}. Recently a huge improvement is done to discriminator models in GAN architecture as an effort to train more effective generator models. 
\par Nvidia’s research team published Style Generative Adversarial Network also known as StyleGAN~\cite{stylegan} as an extension to the existing GAN architecture that proposes large changes to the generator model. This includes the use of a mapping network to map points in latent space to an intermediate latent space. Secondly, they have focused on using the intermediate latent space to control style at each point in the generator model, and finally the introduction to noise as a source of variation at each point in the generator model. This in turn makes the proposed architecture capable of generating high-quality photo-realistic synthetic data. It also provides control over the style of the generated image at different levels of detail by changing the style vectors and noise level. Based on the impressive results of StyleGAN in data-driven unconditional generative image modeling, a newer version of StyleGAN referred to as StyleGAN2~\cite{stylegan2} was released with further improvements.  This includes restructuring adaptive instance normalization to avoid droplet artifacts. Adaptive instance normalization~\cite{AdapInsNor} is a normalization layer that helps in achieving faster and more realistic neural style transfer.  

\par We can find studies where researchers have used StyleGAN and StyleGAN2 by adopting transfer learning methodology for generating artificial human facial data and face attributes\cite{Thermal,yin2022styleheat,kusunose2022facial,song2021agilegan}. 
Moreover, StyleGAN3~\cite{stylegan3} has been recently published, which is mainly used to improve image rotation and image translation, however, the training process is time-consuming, and its latent space is more entangled than its predecessors~\cite{AlalufThird} therefore in this work, we have mainly focused on using StyleGAN2 for rendering synthesized child data. 
\section{Building ChildGAN using StyleGAN} \label{sec3}
\begin{figure*}[!htb]
\begin{center}
\includegraphics[width=\linewidth,height=7.5cm]{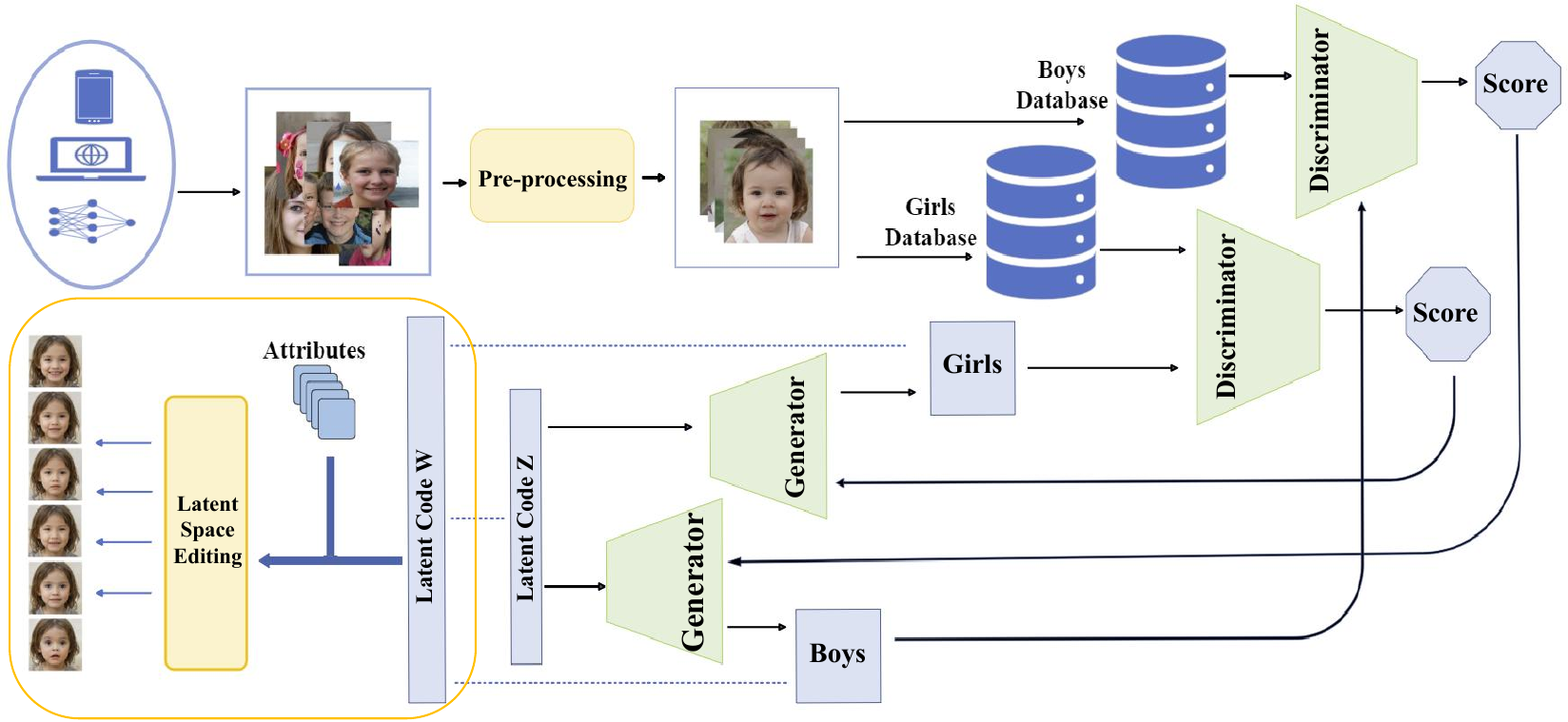}
\end{center}
   \caption{Workflow of the proposed ChildGAN architecture.}
\label{fig_2}
\end{figure*}

In this section, the proposed algorithm is introduced. Our goal is to train a network that can generate photo-realistic boys and girls facial samples separately with controlled attributes. The pipeline of this framework consists of three components as shown in Figure~\ref{fig_2}. This includes, collecting synthetic data for seed training datasets. Secondly, training boys and girls models separately via the ChildGAN generator. Finally, generating faces with different attributes such as expressions, lighting, and ages.

\begin{figure}[htb]
\begin{center}
\includegraphics[width=\linewidth]{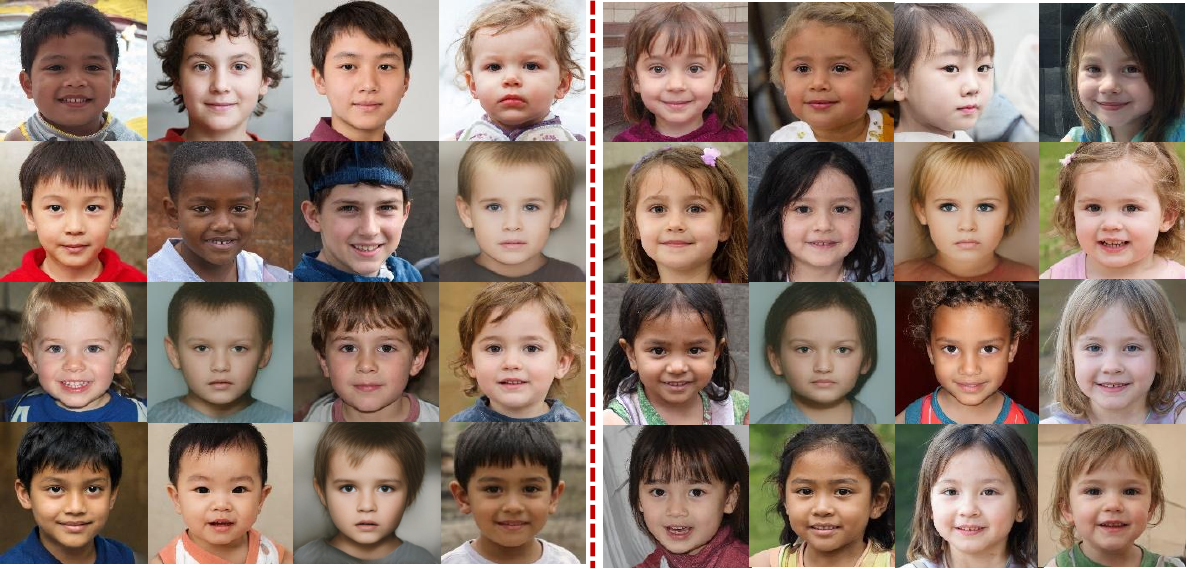}
\end{center}
   \caption{Samples of acquired training data, the left-hand side shows the boys’ data, and the right-hand shows the girls’ data.}
\label{fig_4}
\end{figure}

\subsection{Dataset collection and cleaning}
In this paper, the seed training data 
is all synthetic faces, which are collected from multiple sources. This includes various android applications such as baby generator, fake face generator, and GANs networks such as pretrained StyleGAN and StyleGAN2 as explained in the supplementary material. 
It is important to ensure that all the data samples contain clear and distinguishable facial region of interest. Therefore, the acquired data undergoes certain filtering processes essential for the optimal training of GANs. This incorporates shortlisting the samples which are of high quality, selecting distinct samples, and having no artifacts. 
The overall training data is divided into binary classes i.e., boys and girls with equal distribution to avoid imbalanced dataset issues. In addition, the acquired data is cropped and resized to $1024 \times 1024$, which is the initial size selected for training purposes. Some samples from the training dataset are shown in Figure~\ref{fig_4}.

\subsection{StyleGAN2 training methodology to build ChildGAN }
Motivated by the excellent domain transfer capability, and further disentangled and hierarchical noise introduction in StyleGAN2 architecture we used this network to build ChildGAN by adapting transfer learning methodology. This eventually allows us to smoothly learn new feature vectors for generating high-quality artificial child face samples and further rendering the diversified attributes of the generated children by controlling the latent code. The generation process works by first producing latent code $z$. Subsequently, $z$ is mapped to the intermediate latent space $\mathcal{W}$ to obtain the transformed $w$ for disentanglement. Then, $w$ is converted to style space through affine transformation, which is used to generate synthetic child data. Meanwhile, noise is added to control its details. The discriminator structure employs the ProGAN network~\cite{karras2017progressive}. It is an extension of the training process of GAN that helps the generator models to train with stability which can produce high-quality images. For tuning ChildGAN networks we have used the pre-trained models trained on the $1024 \times 1024$ FFHQ dataset~\cite{stylegan} and made available by NVIDIA in their public Github repositories~\cite{premodel}. The overall training time during each experiment took approximately $3$ days in adversarial setting for generating the outputs with an image size of $1024 \times 1024$.
\subsection{Latent space editing}\label{latentedit}
\noindent
\textbf{Acquiring latent code:} In order to generate smart transformations, the latent codes of the input faces should be acquired before editing the faces. This can be achieved by using two different approaches. The first approach is using the pretrained model to generate optimal result/face and preserve the original latent code $z \in \mathcal{Z}$. Then convert $z$ to easy-to-manipulate intermediate latent code $w$ by learning a mapping $f:z \rightarrow w$, where $w \in \mathcal{W}$ denotes the $(18, 512)$ dimensional latent vectors. The second way is choosing an appropriate GAN inversion method to encode the existing children’s faces. It shows that distortion and editability are a trade-off~\cite{e4encoder} in designing an encoder for StyleGAN image manipulation. In this work e4e encoder~\cite{e4encoder} has been selected to get better editability results. The fitted latent vector $w^*=e4e(x)$ is close to the original $w$ after optimization. The complete algorithm for generating various smart image transformations on child data using latent code is detailed in supplementary material. 

\noindent
\textbf{Facial attributes edit:} The goal of face attribute editing is to allow the modification of one attribute while maintaining other attributes in a similar face. Due to the entanglement of various attributes, it is still challenging to accomplish this task. Studies~\cite{StyleSpace} indicate that style latent code $s$ and intermediate latent code $w$ could perform better attribute editing. In this work, we have focused to learn more diversified and distinct semantical attribute directions such as age and expression corresponding to the latent space $\mathcal{W}$. For this goal, multiple faces have been acquired from our trained model, and then we classify these faces for the target attributes. Microsoft Cognitive Services API~\cite{api} has been used for face recognition to obtain the attribute labels of faces. Then logistic regression classifier has been used for training purpose to get the semantical attribute directions $\alpha_1, \alpha_2, ..., \alpha_n$. Thus, the new face with the corresponding attribute is elaborated in equation~\ref{eq2}.
\begin{equation}\label{eq2}
w_{update} = w+\alpha_t \cdot \mathit{coeff}
\end{equation}
Where $w_{update}$ is the updated latent codes, $\alpha_t$ is the semantical direction, and $\mathit{coeff}$ is the adjustable vector to control the new attribute. Finally, the updated face image $I_{update}$ is generated as shown in equation~\ref{eq3}. 
\begin{equation}\label{eq3}
I_{update} = Generator(w_{update},\theta)
\end{equation}
Where $\theta$ are the parameters of the generator.

\noindent
\textbf{Relighting portrait :} For embedding efficient lighting effects on synthetic child facial data, we have explored several SOTA algorithms and shortlisted Deep Single-Image Portrait Relighting (DPR) technique~\cite{DPR} for this work.
 The adapted technique has shown robust qualitative and quantitative results on various datasets~\cite{DPR}. Moreover, this method is effective as it particularly avoids artifacts due to the physically based relighting method as compared to other re-lighting techniques~\cite{sfsnet, 10.1145/3414685.3417824}.
This method adopts the spherical harmonics lighting model which allows more precise control of directional lighting as compared to latent editing for lighting control. The pretrained model originally trained on the CelebA-HQ dataset~\cite{karras2017progressive} has been used for producing more than 60  distinctive face lighting conditions thus illuminating different facial regions. 

\section{Experimental results}\label{sec4}
This section will present the tuned StyleGAN2 outputs along with various smart transformations. 
Supplementary material provides further details of machine configuration along with the optimal set of shortlisted network hyperparameters for robust tuning of pre-trained StyleGAN2 networks and training results in the form loss graphs.

\subsection{Model evaluation using quality metrics}
\begin{figure}
\begin{center}
\includegraphics[width=0.85\linewidth]{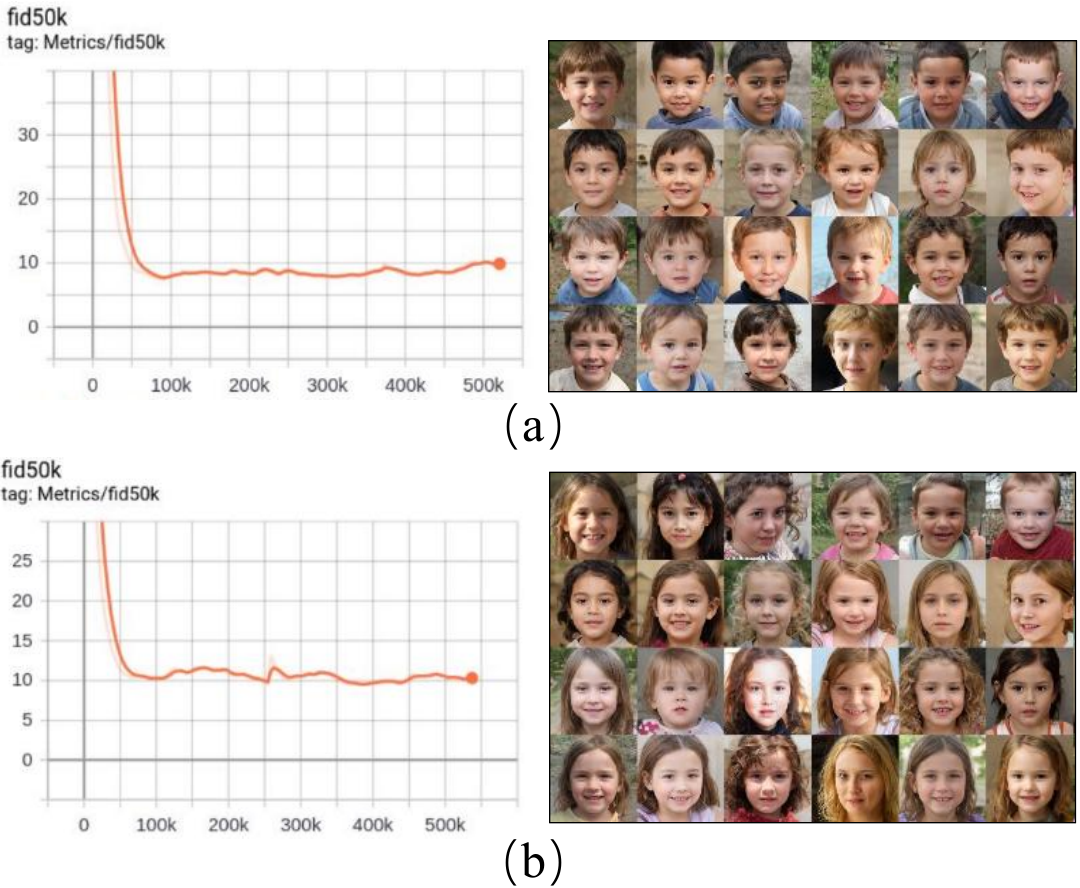}
\end{center}
   \caption{FID metric results with images generated at the final epoch of ChildGAN architecture, a) shows the FID results for the boys model, b) shows the FID results for the girls model.}
\label{fig_7}
\end{figure}

After the training process using transfer learning methodology, the next stage includes the validation performance of optimally tuned models. 
Figure~\ref{fig_7} illustrates the tuned model evaluation results by incorporating Fréchet Inception Distance score (FID) metrics~\cite{fid}. This metric is generally used to estimate the distance between feature vectors calculated for real and generated images for trained GAN architectures as shown in equation~\ref{eq4}.
\begin{equation}\label{eq4}
FID(x,g)=||\mu_x-\mu_g||+Tr(\Sigma_x-\Sigma_g-2\sqrt{\Sigma_x\Sigma_g})
\end{equation}
Where $x$ and $g$ are the real and generated embeddings. $\mu_x$ and $\mu_g$ are the magnitudes of $x$ and $g$. $Tr$ is the trace of the matrix and $\Sigma_x$ and $\Sigma_g$ are the covariance matrix. 
As can be observed from Figure~\ref{fig_7} the FID curve for both the tuned models is getting lower which means the generated data has better image quality and diversity. 
\subsection{Child transformation results using latent code and relighting algorithm}
This section will discuss various child transformation outputs as explained in section~\ref{latentedit}. In this work, we have modified the synthetic child facial outputs by adapting 5 distinct smart transformations. These transformations include rendering faces with natural facial expressions, eye blinking effect, hair and skin color digitization, child aging, and head pose variation. In addition to that we have incorporated different lighting  conditions using DPR method based on deep learning to illuminate facial regions to further bring in data diversity.

\begin{figure}[htb]
\begin{center}
\includegraphics[width=0.9\linewidth]{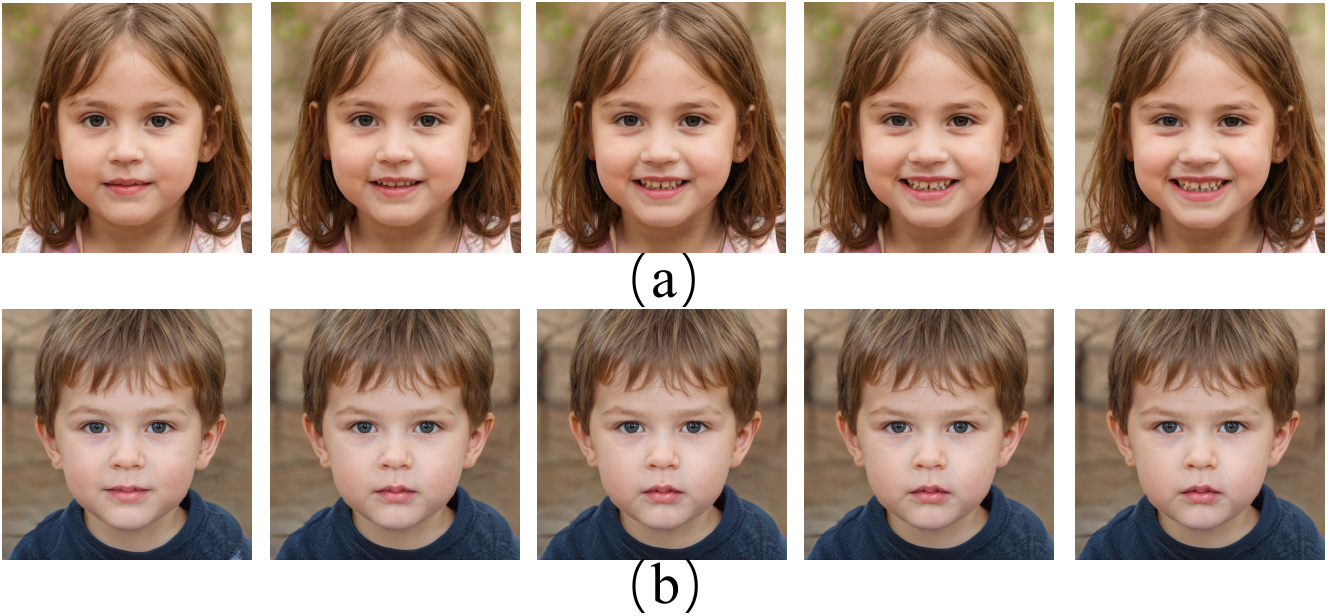}
\includegraphics[width=0.9\linewidth]{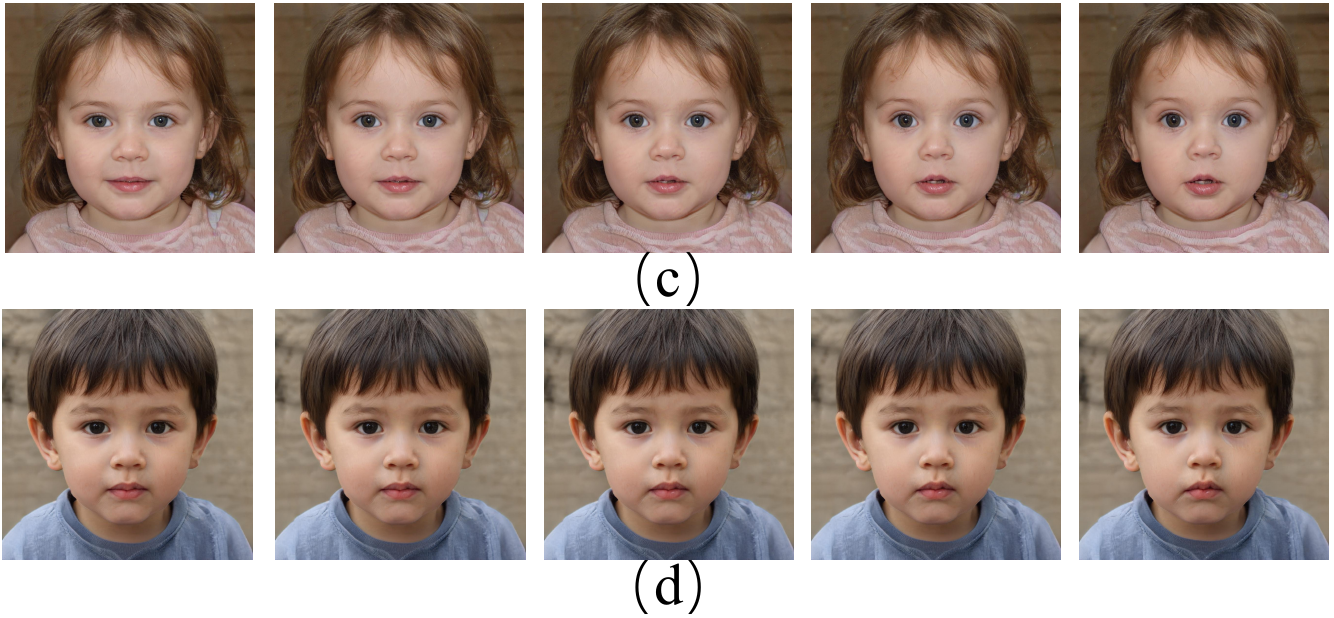}
\end{center}
   \caption{Child facial expression results on four different subjects acquired from synthetic boys and girls data, a) shows neutral to happy expressions, b) shows neutral to angry expressions, c) shows neutral to surprise expressions, and d) shows neutral to sad expressions.}
\label{fig_8}
\end{figure}

\noindent
\textbf{1. Facial expressions:} A human face has significant and distinctive characteristics that help in the recognition of facial expressions caused by an individual's internal emotional state. Facial expression analysis is used in a wide range of human-computer interaction (HCI) applications, such as face image processing, and facial animation~\cite{PiseApplication}. In this work, four different facial expressions are generated using w latent code and stored as shown in Figure~\ref{fig_2}. These expressions include neutral to happy, neutral to angry, neutral to surprise, and neutral to sad as depicted in Figure~\ref{fig_8}

\begin{figure}[htb]
\begin{center}
\includegraphics[width=0.9\linewidth]{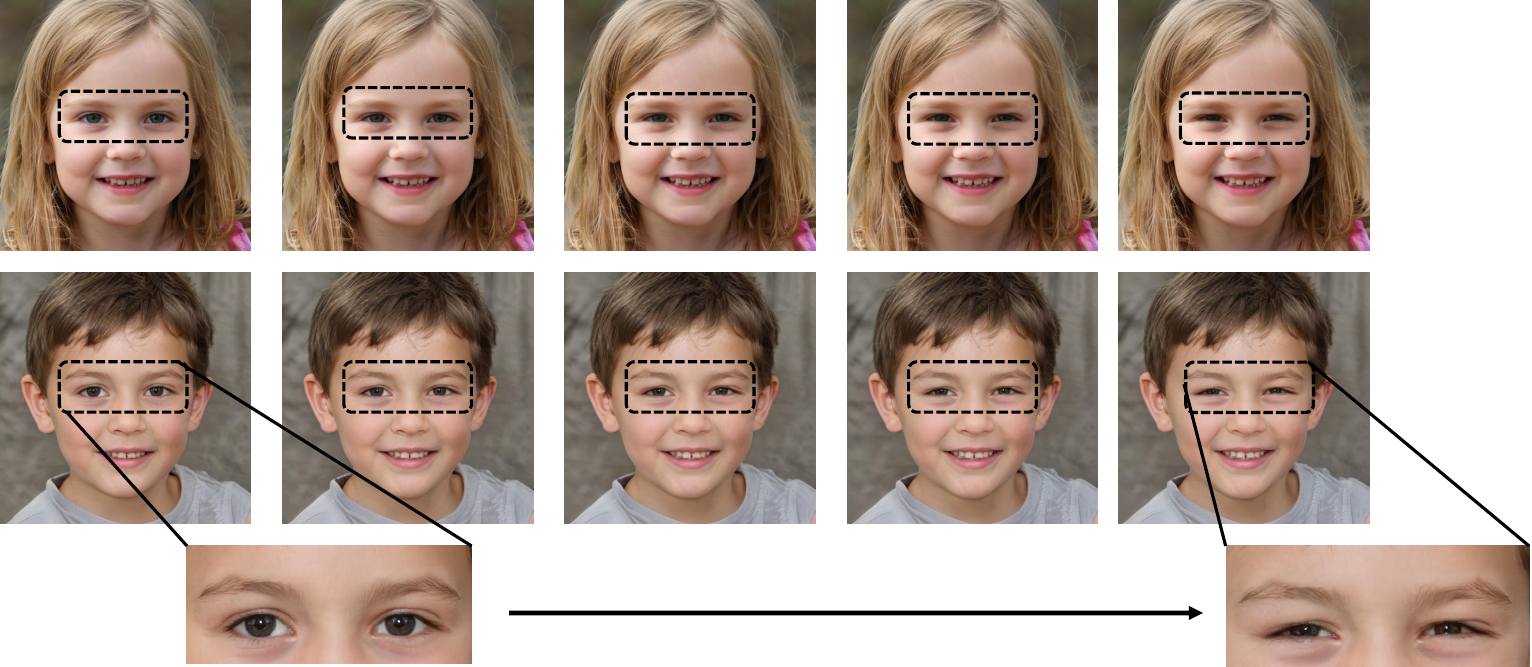}
\end{center}
   \caption{Synthetic child facial eye blinking results on two different subjects.}
\label{fig_9}
\end{figure}

\noindent
\textbf{2. Eye blinking:} The eye blinking effect can be applied in certain child facial analysis tasks, including anti-spoofing and drowsiness detection and may help detect some health disorders. In this work, we have applied the eye-blinking effect in different lighting conditions. This is achieved through latent code by adjusting the attribute directions and loading the tuned model architecture. This allows to develop blink detection algorithms that are more robust to environments with uncontrolled lighting conditions. Figure~\ref{fig_9} shows the eye blinking results of two different subjects.

\begin{figure}[htb]
\begin{center}
\includegraphics[width=0.8\linewidth]{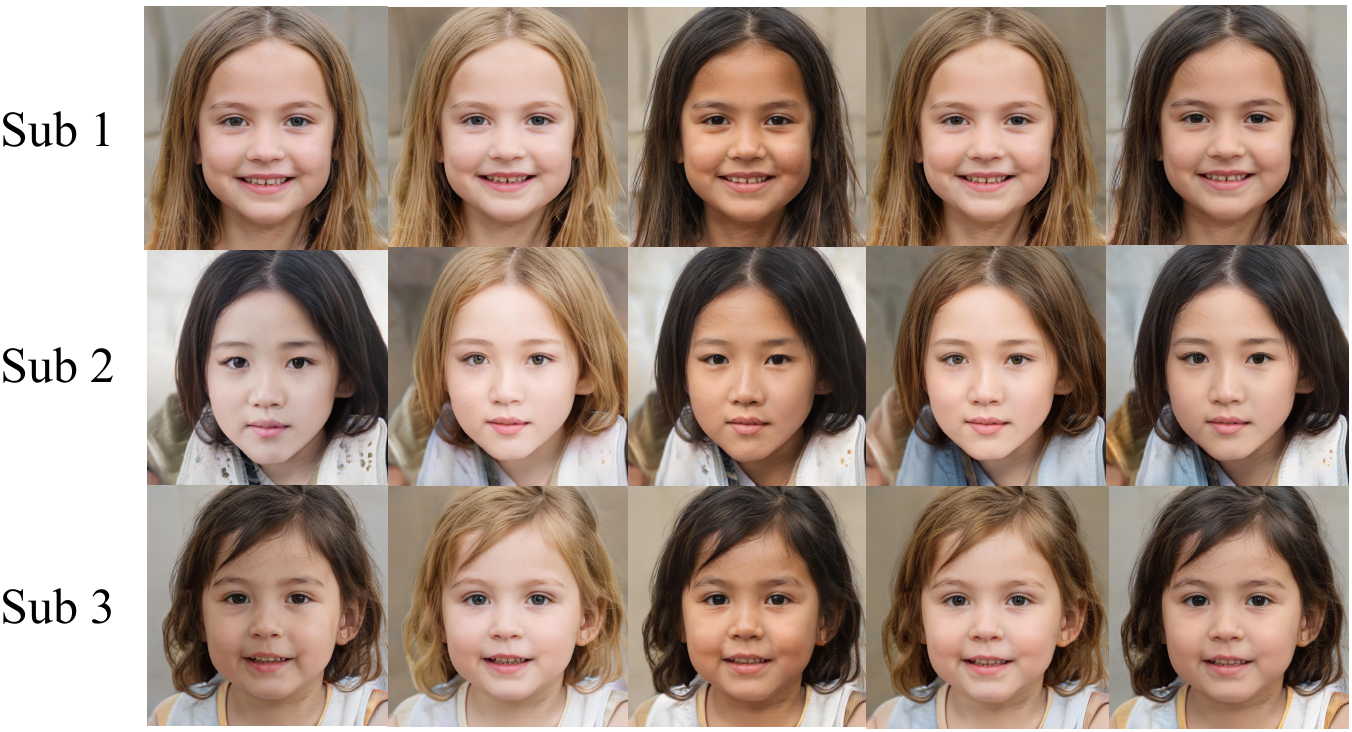}
\end{center}
   \caption{Synthetic child facial skin tone and hair color transformation on three different subjects.}
\label{fig_10}
\end{figure}

\noindent
\textbf{3. Skin tone and hair color digitization:} The next smart transformation includes further rendering the generated child data with distinct skin tones and hair colors.  This type of transformation is useful for developing automated systems for capturing the appearance of a physical hair sample, skin tone-based dress, and makeup selection for children and for augmented/virtual reality applications. This is accomplished by mixing different latent codes to get unique skin color adaptations and hair colors. Figure~\ref{fig_10} shows the generated effects on girls’ data.

\begin{figure}[htb]
\begin{center}
\includegraphics[width=\linewidth]{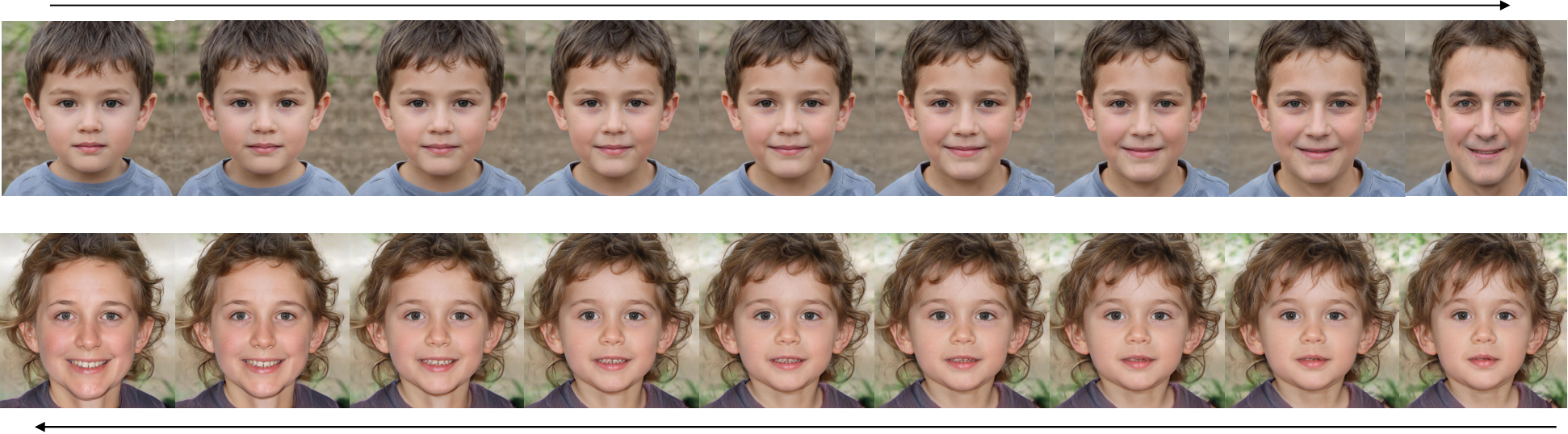}
\end{center}
   \caption{Synthetic child aging progression results, the first row shows the aging progression results of a boy from child to young \& the second row shows the aging progress from young to the child for a girl.}
\label{fig_11}
\end{figure}

\noindent
\textbf{4. Child aging augmentation:} Age progression is a challenging factor for face recognition algorithms deployed in smart mobile and embedded platforms. The complication is primarily based on by biological changes that occur during aging and can result in visible changes in facial characteristics when analyzing the images of the same person taken at different ages. There is an emerging need to extract robust facial features for age-invariant face recognition systems because the face is part of the human body that is greatly affected by aging. This section will detail the experimental outcomes of child aging transformation results using latent code. The results were generated on more than 500 different boys and girls subjects. Figure~\ref{fig_11} shows the child aging transformation results on two separate subjects.

\begin{figure}[htb]
\begin{center}
\includegraphics[width=\linewidth]{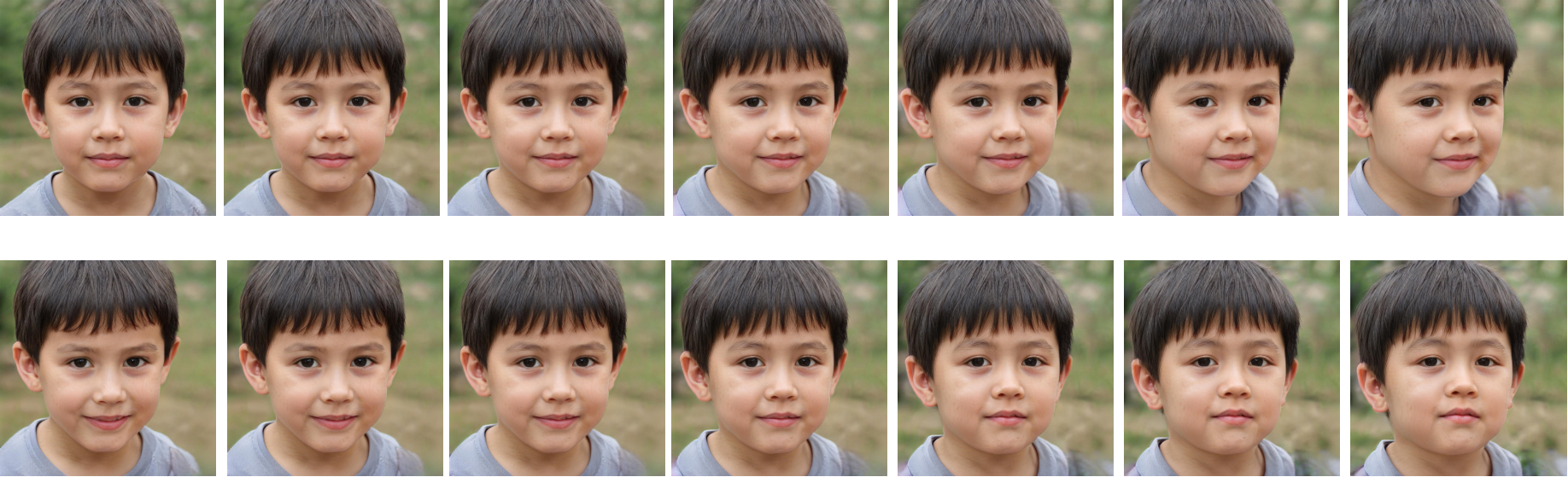}
\end{center}
   \caption{Synthetic child head pose variation results, the first row shows the yaw variation, and the second row shows the pitch variation.}
\label{fig_12}
\end{figure}

\noindent
\textbf{5. Face head pose variation:} In addition to facial expression, eye blinking, skin and hair color variation, and aging progression transformation we have also included face head pose variation using latent code.  Generating children’s head pose data is beneficial for various computer vision applications such as aiding in gaze estimation, modeling attention, fitting 3D models to video, and performing face alignment. Figure~\ref{fig_12} shows the child’s facial head pose variation results.

\noindent
\textbf{6. Relighting face:} Human faces can have a lot of variations in appearance due to distinct lighting conditions which eventually becomes a challenge for face recognition algorithms. In this work, we have addressed this barrier by generating more than sixty  different illumination conditions using the DPR method.
Moreover, for this work, we have further shortlisted the four most distinctive lighting
conditions covering four directional facial regions i.e up, down, left, and right. The details along with all the lighting condition results are
explained in the supplementary material. Figure~\ref{fig_14} depicts the selected set of lighting conditions along with facial illumination results covering different facial angles.


\begin{figure}[htb]
\begin{center}
\includegraphics[width=\linewidth]{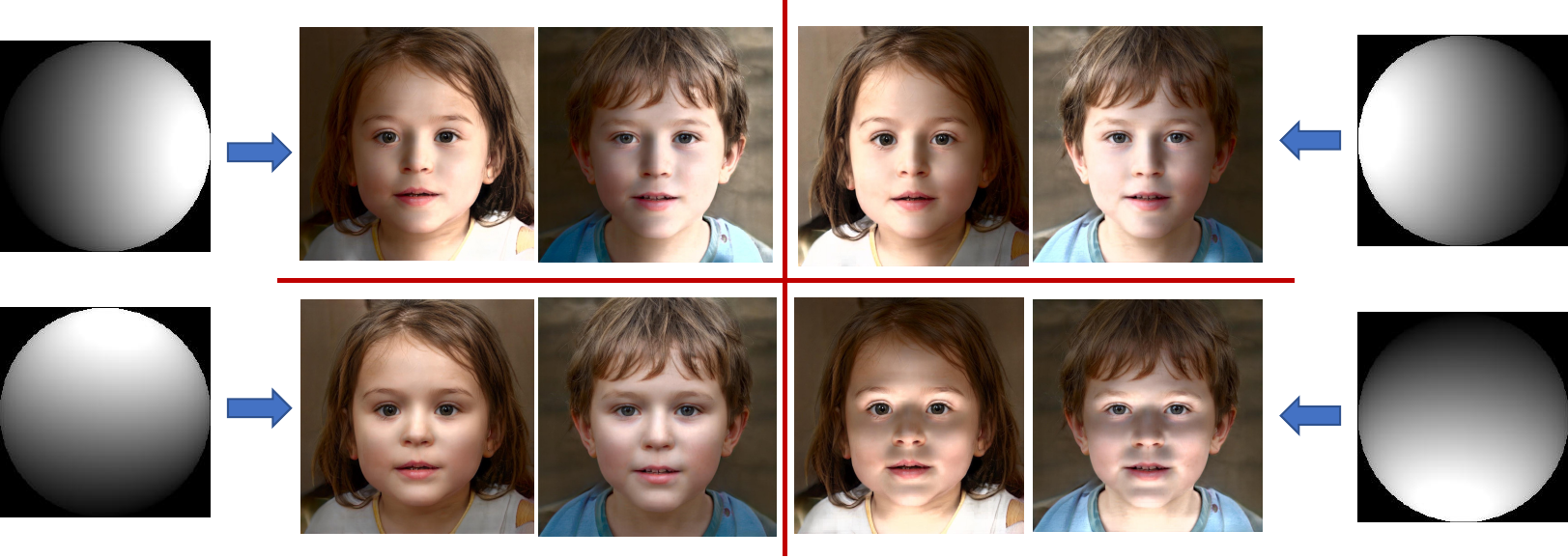}
\end{center}
   \caption{Synthetic child face data with four distinct lighting conditions.}
\label{fig_14}
\end{figure}

Table~\ref{tab3} shows the overall generated child data along with its attributes using various smart transformations as discussed in section~\ref{sec3} and section~\ref{sec4}. Thus we have generated a total data of around three hundred and twenty four thousand distinct samples of boys and girls using ChildGAN. The additional child data sample results along with the different smart transformations can be found in the supplementary material. The complete dataset will be open-sourced via github page and can be used for commercial and non-commercial applications.

\section{ChildGAN data validation}\label{sec5}
With the rapid advancement of machine learning algorithms and data required for real-world computer vision applications, the lack of high-quality data is the real bottleneck in the AI industry.  This section will focus on different computer vision and machine learning techniques to check the quality of generated child synthetic data. The details of each of these validation methods are provided in the below subsections.
\begin{table}[htb]
\begin{center}
\resizebox{0.9\linewidth}{!}{
\begin{tabular}{|l|l|l|l|}
\hline 
Category &
  \begin{tabular}[c]{@{}c@{}}No. of\\ Sam.\end{tabular} &
  \begin{tabular}[c]{@{}c@{}}No. of\\ Sam./Subj.\end{tabular} &
  Total \\ \hline \hline
Child Data &
  \begin{tabular}[c]{@{}l@{}}Boys: 10k\\ Girls: 10k\end{tabular} &
  1 &
  20k \\ \hline
Facial Expressions &
  \begin{tabular}[c]{@{}l@{}}Boys: 2k\\ Girls: 2k\end{tabular} &
  \begin{tabular}[c]{@{}l@{}}Happy: 6\\ Sad: 6\\ Angry: 6\\ Surprise: 6\end{tabular} &
  96k \\ \hline
Eye blinking effect &
  \begin{tabular}[c]{@{}l@{}}Boys: 2k\\ Girls: 2k\end{tabular} &
  6 &
  24k \\ \hline
\begin{tabular}[c]{@{}l@{}}Skin Tone and Hair\\ Color Variation\end{tabular} &
  \begin{tabular}[c]{@{}l@{}}Boys: 1k\\ Girls: 1k\end{tabular} &
  6 &
  12k \\ \hline
Aging Progression &
  \begin{tabular}[c]{@{}l@{}}Boys: 2k\\ Girls: 2k\end{tabular} &
  8 &
  32k \\ \hline
Headpose Variation &
  \begin{tabular}[c]{@{}l@{}}Boys: 2k\\ Girls: 2k\end{tabular} &
  \begin{tabular}[c]{@{}l@{}}Yaw: 8\\ Pitch: 7\end{tabular} &
  60k \\ \hline
Relighting &
  \begin{tabular}[c]{@{}l@{}}Boys: 10k\\ Girls: 10k\end{tabular} &
  4 &
  80k \\ \hline 
\end{tabular}
}
\end{center}
\caption{Overall child synthetic dataset}
\label{tab3}
\end{table}
\subsection{Gender classification test}
\begin{figure}[htb]
\begin{center}
\includegraphics[width=0.85\linewidth]{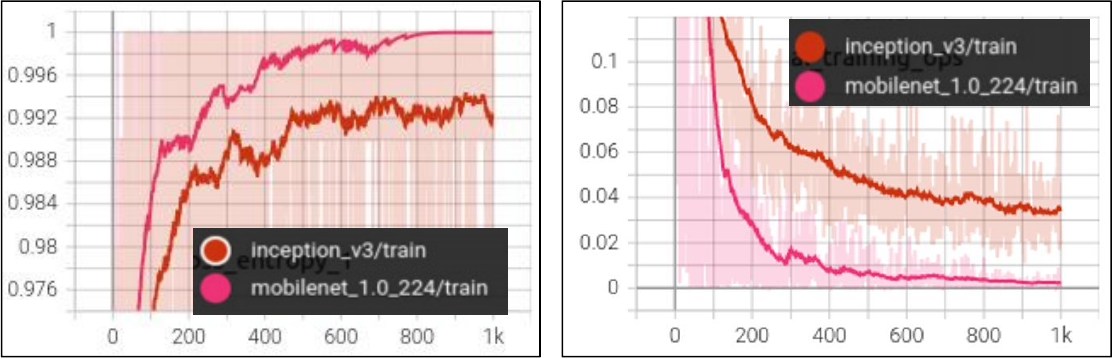}
\end{center}
   \caption{Binary gender classifier training results on child synthetic data, a) shows the accuracy graph of mobileNet-v2 and Inception-v3 CNN with an overall accuracy of 99.87\% and 99.15\%, b) shows the loss graph of mobileNet-v2 and Inception-v3 CNN with a loss rate of 0.015 and 0.061.}
\label{fig_15}
\end{figure}

In this test, we have trained a binary classifier (boys and girls) using end-to-end pre-trained CNN which includes InceptionV3~\cite{inception} and MobileNetV2~\cite{MobileNetV2} on synthetic data. The performance of the trained classifier on synthetic data is then cross-validated on real child data which is acquired from UTKFace dataset~\cite{utkface}. Figure~\ref{fig_15} shows the training and validation results in the form of accuracy and loss graphs. The model was trained for 1000 epochs by selecting binary cross entropy as the loss function and SGD optimizer. The tuned gender classifier models on synthetic data were then validated on a set of 500 distinct test samples. MobileNetV2 and InceptionV3 models achieve overall test accuracy of 94\% and 92\% respectively thus showing the precise robustness of the trained model on synthetic data.
\subsection{Face localization and facial landmark detection test}
\begin{figure}[htb]
\begin{center}
\includegraphics[width=0.9\linewidth]{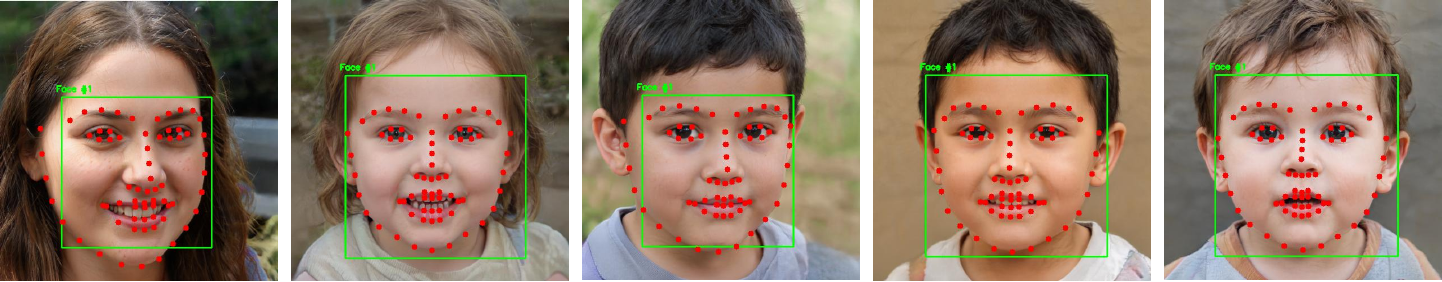}
\end{center}
   \caption{Face mapping and 68 facial landmark detection test results on five different subjects (boys \& girls) with varying face head pose, aging difference, skin tone variation, and different facial expressions.}
\label{fig_16}
\end{figure}

The second critical test includes face localization and facial landmark detection on child synthetic data using DLIB~\cite{dlib} library. The system works by employing a 68-facial-landmark detector to extract multiple facial key points including the child’s eyes, nose, lip, and around the face. This plays a vital role in different computer vision applications such as face recognition, biometrics, expression analysis, etc. The test was performed on 100 distinct samples with varying face poses, different genders, and other smart transformations mentioned in Table~\ref{tab3}. Figure~\ref{fig_16} shows some of the challenging test samples with robust face mapping and face landmark detection results.

\subsection{Synthetic data uniqueness test}
\begin{figure}[htb]
\begin{center}
\includegraphics[width=\linewidth]{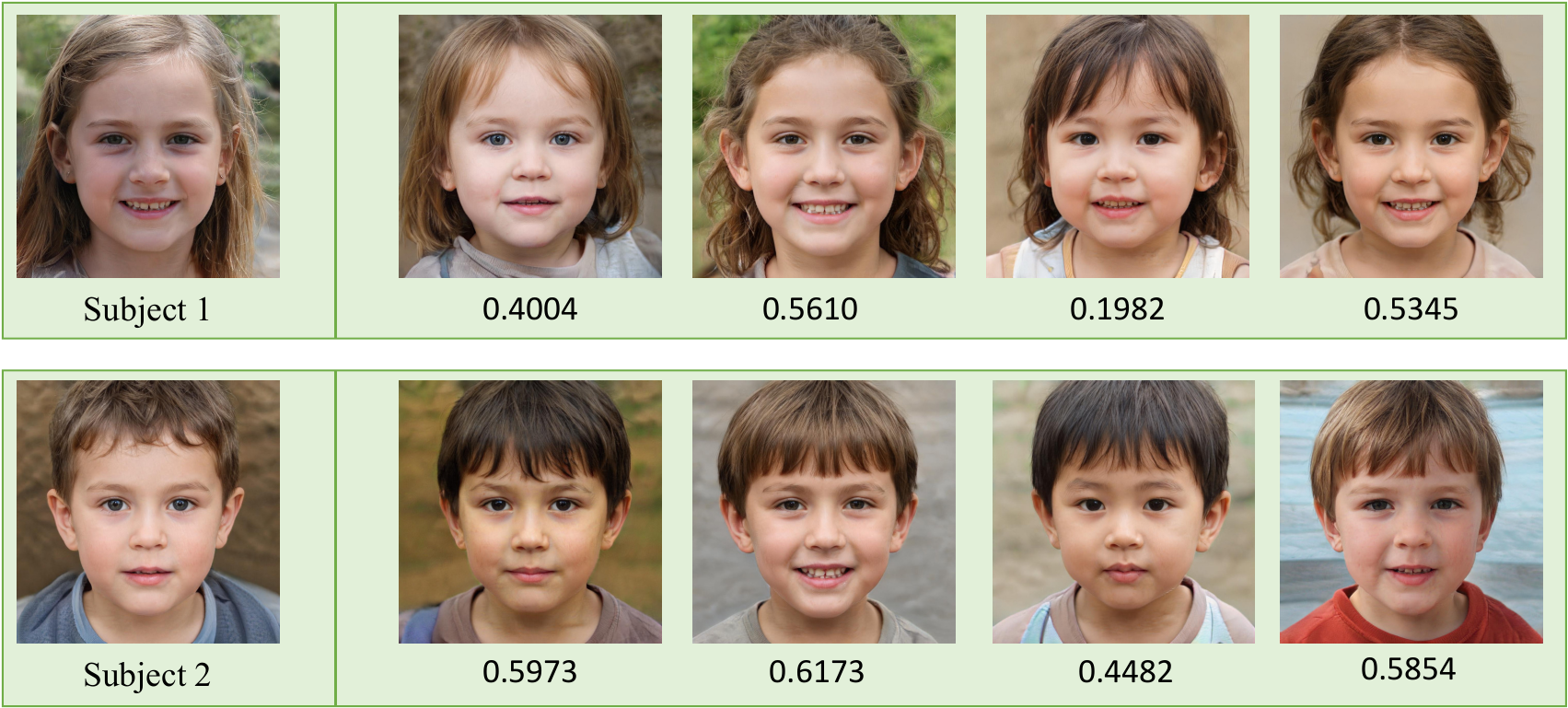}\\
\includegraphics[width=\linewidth]{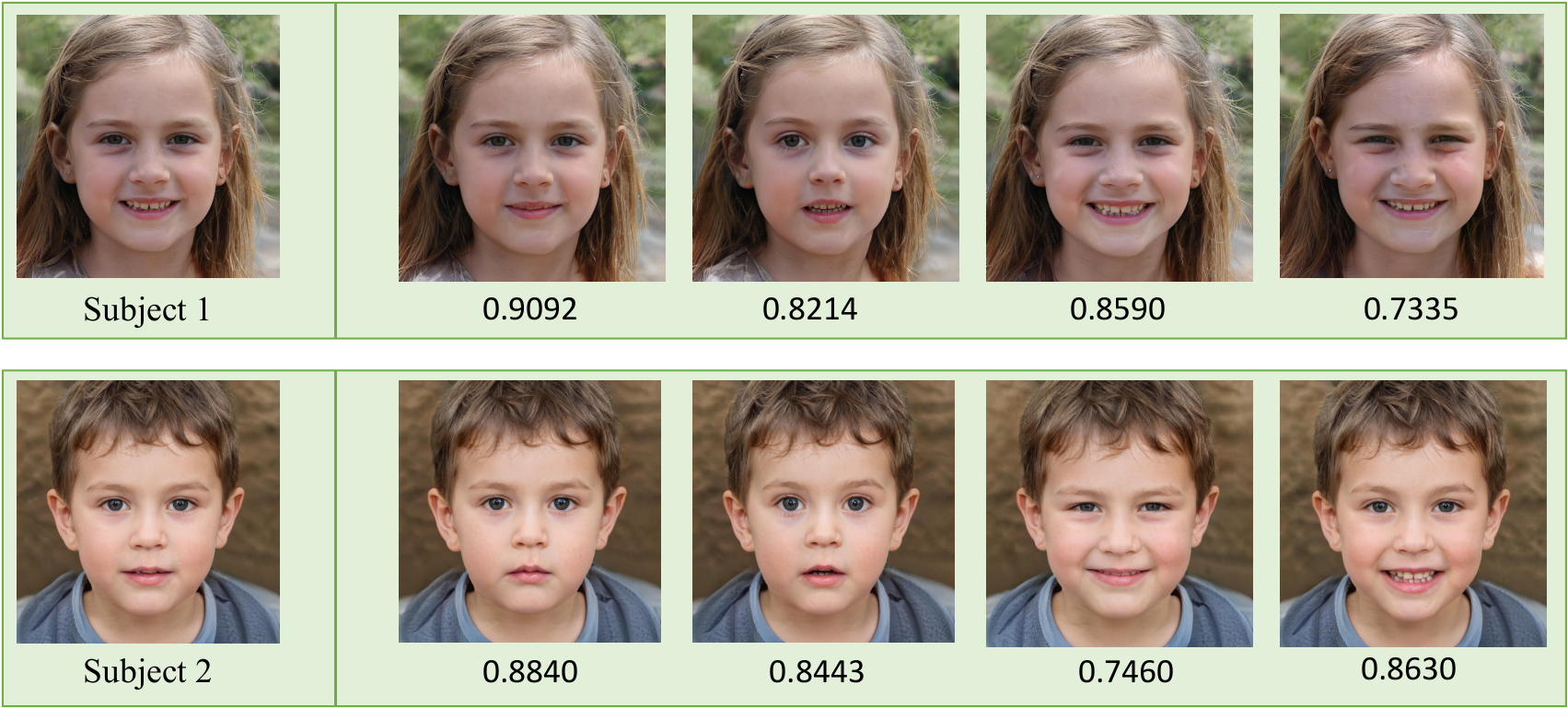}
\end{center}
   \caption{The similarity score between different samples. The first and second row calculates the similarity score of subject 1 and subject 2 with other girls and boys data samples. The third and fourth row shows the similarity score of subject 1 and subject 2 with various expressions of subject 1 and subject 2.}
\label{fig_17}
\end{figure}

In this experiment, we adopt ArcFace, a well-known reference face recognition models ~\cite{arcface} to test the identity similarity of synthetic child faces. The model is used to test the data using two different methods. 
In the first test, we perform the similarity measure between different subjects. Secondly, we test  different facial attributes of the same subject to validate that the identity remains consistent. 
Figure~\ref{fig_17} shows that the cosine similarity of two different subjects is lower than the cosine similarity of other boys and girls samples along with distinct facial expressions of similar subjects. This test demonstrates the facial patterns and features, of synthetic child faces provide useful uniqueness, diversity, and quality.
Additional validation results are provided in the supplemental material.

\subsection{Eye aspect ratio test}
The Eye Aspect Ratio also referred to as EAR, is a scalar value that particularly represents the opening and closing ratio of the eyes~\cite{electronics11193183} using facial landmark detectors across the eyes. During the eye flashing process, the EAR value increases or decreases significantly thus displaying the eye blinking effect. The details of calculating the EAR using eye landmarks is explained in the supplementary material.

\begin{figure}[htb]
\begin{center}
\includegraphics[width=0.6\linewidth]{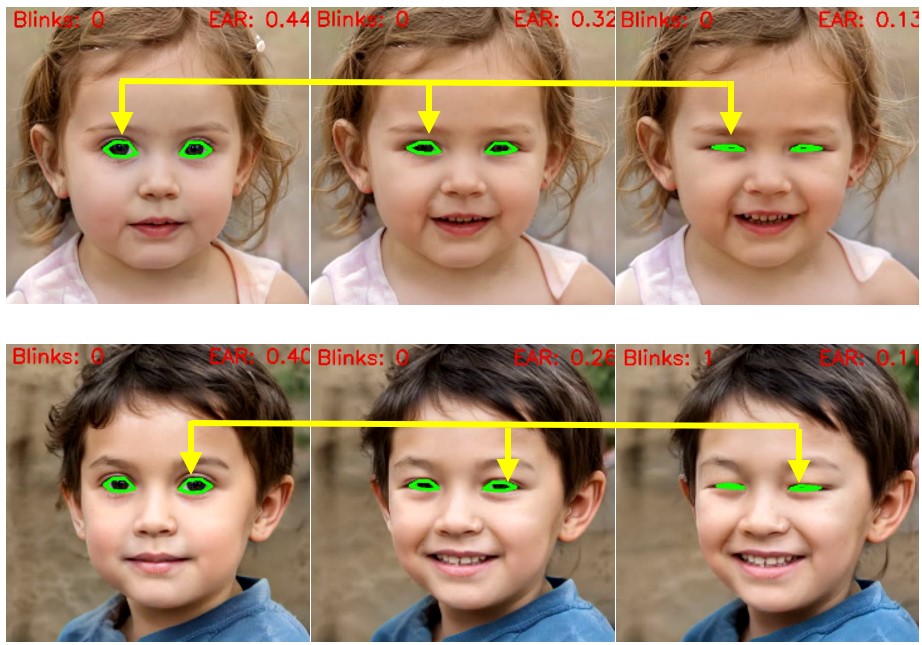}
\end{center}
\caption{Eye blink detection and EAR results. First row shows the girls’ images with the eye blinking effect thus changing EAR from 0.44 to 0.12 whereas the second row shows the boys’ images with the eye blinking effect thus changing EAR from 0.41 to 0.11.}
\label{fig_19}
\end{figure}

In this work, we have employed the EAR test to validate the eye blinking effect quantitatively on synthetic child facial data. Figure~\ref{fig_19} demonstrates eye blinking effect and EAR results by using facial landmarks across the eyes on two different subjects generated through ChildGAN.
By observing Figure~\ref{fig_19} it can be analyzed that EAR value changes significantly during eye blinking which proves that generated child data has photo-realistic facial features.

\section{Conclusion and future work}\label{sec6}
In this work, we have proposed ChildGAN architecture to render synthetic child facial data samples at scale. In addition various smart transformations enable the base data to be modified to increase the variety of data samples for each individual child. The tuned model show that transfer learning in GAN is indeed an effective approach for producing high-quality
images even when the source and target domains are different
from each other. The associated fine-tuned models and methodological details are provided to allow other researchers to build on this work. For now, we have tested the potential of this data by running four different computer vision application tests to validate the qualitative behaviour of this synthetic data. However it remains a challenging task to validate synthetic data distributions as a ground truth (real data distribution) is not readily available to make a direct comparative analysis.

\par Several important research tasks remain as future challenges. 
Firstly, to explore robust qualitative and quantitative metrics to validate the distribution of synthetic identities as we do not have a corresponding set of real data \cite{varkarakis2020validating}. 
The second challenge is to improve the capability of the synthetic data samples to support broader diversity in facial analysis algorithms. This could involve adopting a similar methodology to build ethnicity into the style-transfer and explore if we can maintain a similar identity distribution for children of a specific ethnicity. 
Finally, the potential of this high-quality dataset can be  explored for a variety of 3D computer applications through the reconstruction of 3D child facial models \cite{basak20223d}. It will be interesting to see how the research community builds on this first large-scale  dataset of synthetic child facial data.






{\small
\bibliographystyle{ieee_fullname}
\bibliography{childgan}
}

\end{document}